\documentclass{article}

\usepackage{iclr2027_conference_based_arxiv,times}
\usepackage{wrapfig}
\usepackage{multirow}
\usepackage{makecell}

\usepackage{amsmath,amsfonts,bm}

\def\eqref#1{equation~\ref{#1}}

\def\1{\bm{1}}

\DeclareMathAlphabet{\mathsfit}{\encodingdefault}{\sfdefault}{m}{sl}
\SetMathAlphabet{\mathsfit}{bold}{\encodingdefault}{\sfdefault}{bx}{n}

\usepackage{graphicx}
\usepackage{wrapfig}
\usepackage{booktabs}
\usepackage{array}
\usepackage{hyperref}
\usepackage{url}
\usepackage[capitalize]{cleveref}

\crefname{section}{Sec.}{Secs.}
\Crefname{section}{Sec.}{Secs.}
\crefname{figure}{Fig.}{Figs.}
\Crefname{figure}{Fig.}{Figs.}
\crefname{table}{Tab.}{Tabs.}
\Crefname{table}{Tab.}{Tabs.}

\graphicspath{{figures/}}

\providecommand{\bestresult}[1]{{\setlength{\fboxsep}{1.5pt}\colorbox[rgb]{0.78,0.94,0.68}{\strut #1}}}
\title{WorldAgent: Verification-Guided Agentic\\Physical World Construction}

\author{
\textbf{Caoliwen Wang}$^{1,*}$ \quad
\textbf{Mengdi Wang}$^{1,*,\dagger}$ \quad
\textbf{Yige Chen}$^{2,*,\S}$ \quad
\textbf{Zejia Wu}$^{3}$ \quad
\textbf{Bowen Huang}$^{3}$ \\[0.03em]
\textbf{Siyuan Chen}$^{1}$ \quad
\textbf{Guanxiong Chen}$^{1}$ \quad
\textbf{Lifu Wei}$^{1}$ \quad
\textbf{Heng Zhang}$^{1}$ \quad
\textbf{Qinghai Zhang}$^{3}$ \\[0.03em]
\textbf{Yin Yang}$^{4}$ \quad
\textbf{Guandao Yang}$^{5}$ \quad
\textbf{Shiying Xiong}$^{3}$ \quad
\textbf{Peng Wang}$^{2}$ \\[0.03em]
\textbf{Chenfanfu Jiang}$^{6}$ \quad
\textbf{Peter Yichen Chen}$^{1}$ \\[0.03em]
\normalfont
$^{1}$University of British Columbia \quad
$^{2}$VAST \quad
$^{3}$Zhejiang University \quad
$^{4}$University of Utah \\[0.03em]
$^{5}$The University of Texas at Austin \quad
$^{6}$University of California, Los Angeles
}

\iclrfinalcopy

\begin{document}

\maketitle
\begingroup
\renewcommand\thefootnote{\fnsymbol{footnote}}
\footnotetext[1]{Equal contribution.}
\footnotetext[2]{Corresponding author.}
\footnotetext[4]{Work done during internship at VAST.}
\endgroup
\begin{abstract}
Constructing complex physical worlds from language requires coordinating extensive 3D environments, detailed structures and objects at different spatial scales, and interacting physical processes under both stated goals and implicit physical constraints. We present \emph{WorldAgent}, an agentic framework for verification-guided physical world construction from a single natural-language prompt, without iterative user debugging. A world construction layer expands the prompt into a structured world specification and uses physical knowledge to build scenes and run numerical simulations. After every step, a verification layer inspects scene geometry and simulation states alongside rendered views. Failed checks guide automatic revisions to the specification and re-execution of the affected steps. Accepted worlds pass the required checks and remain editable for further inspection and resimulation. We introduce \emph{AgenticSimBench}, on which WorldAgent achieves the best scores among the evaluated agent-based methods on five of seven metrics. In a 26-participant user study, it receives the highest mean ratings across all four criteria.
\end{abstract}

\section{Introduction}
\label{sec:introduction}

AI agents increasingly combine coding and tool use for complex tasks. Can these capabilities extend to designing physical worlds with complex 3D structure and physically accurate behavior? Such worlds could reduce tedious modeling and simulation setup in games and film, and support scientific studies through digital models of natural environments.

Constructing detailed 3D scenes requires translating underspecified descriptions into modeling choices~\citep{sun2023threedgpt} and arranging assets under spatial constraints~\citep{hu2024scenecraft,yang2024holodeck}. Physical worlds additionally require correct behavior: a box resting on a table requires the table's support to balance its weight, while a stream carrying leaves requires water to flow and interact with the leaves. A prompt rarely spells out these physical conditions, even when the intended outcome depends on them. If a system leaves them implicit, the user must inspect each failed result, identify the missing constraint, and clarify the request over multiple turns. This burden grows as more objects and physical processes interact. Numerical simulators make the inferred relationships executable through discretized governing equations, giving agents a way to realize and test physical behavior. World construction therefore involves configuring both a scene and the simulator that governs its evolution.

Simulator execution does not by itself ensure the intended evolution. Nonlinear dynamics and changing contacts can make outcomes sensitive to initial conditions: small changes in a die's release orientation or velocity can alter its final resting face. Inferring a setup for prescribed behavior can become impractical even with differentiable simulation: discontinuities and stiff dynamics can undermine gradient estimates~\citep{suh2022differentiable}, while contact gradients may vanish when the required contact never occurs~\citep{paulus2026hardcontacts}. Agents therefore execute candidate worlds, verify both physical consistency and task fulfillment, and revise configurations using the observed discrepancies. This verification loop provides evidence for both refinement and acceptance.

Recent agentic simulation systems combine simulator execution with feedback-driven scene construction and refinement~\citep{liu2026simworlds,zhang2026gsagent,wang2026chronoagentic}. We present \emph{WorldAgent}, a framework for \emph{verification-guided physical world construction} from a single natural-language prompt. It coordinates extensive environments, detailed local structures, and multiple interacting physical processes through a structured world specification. The world construction layer builds scenes and runs simulations; after each step, the verification layer inspects geometry and simulation states alongside rendered views. This exposes violations such as occluded interpenetration or liquid-volume loss and guides automatic specification revisions and re-execution, producing editable worlds without iterative user debugging. On \emph{AgenticSimBench}, WorldAgent leads the evaluated agent baselines on five of seven metrics using the same language model backbone. A 26-participant user study gives it the highest mean ratings across all four criteria.

We summarize our contributions as follows.
\begin{itemize}
    \item A verification-guided refinement method that checks each construction and simulation step and uses observed failures to revise the structured world specification.
    \item An agentic framework that integrates physical knowledge, scene construction, and numerical simulation to autonomously construct complex, editable physical worlds with rich scene structure and interacting physical processes from a single natural-language prompt.
    \item The design of \emph{AgenticSimBench} and an evaluation protocol for physical consistency, task fulfillment, and component contributions across complex dynamic scenes.
\end{itemize}

\begin{figure*}[t]
    \centering
    \includegraphics[width=0.94\textwidth]{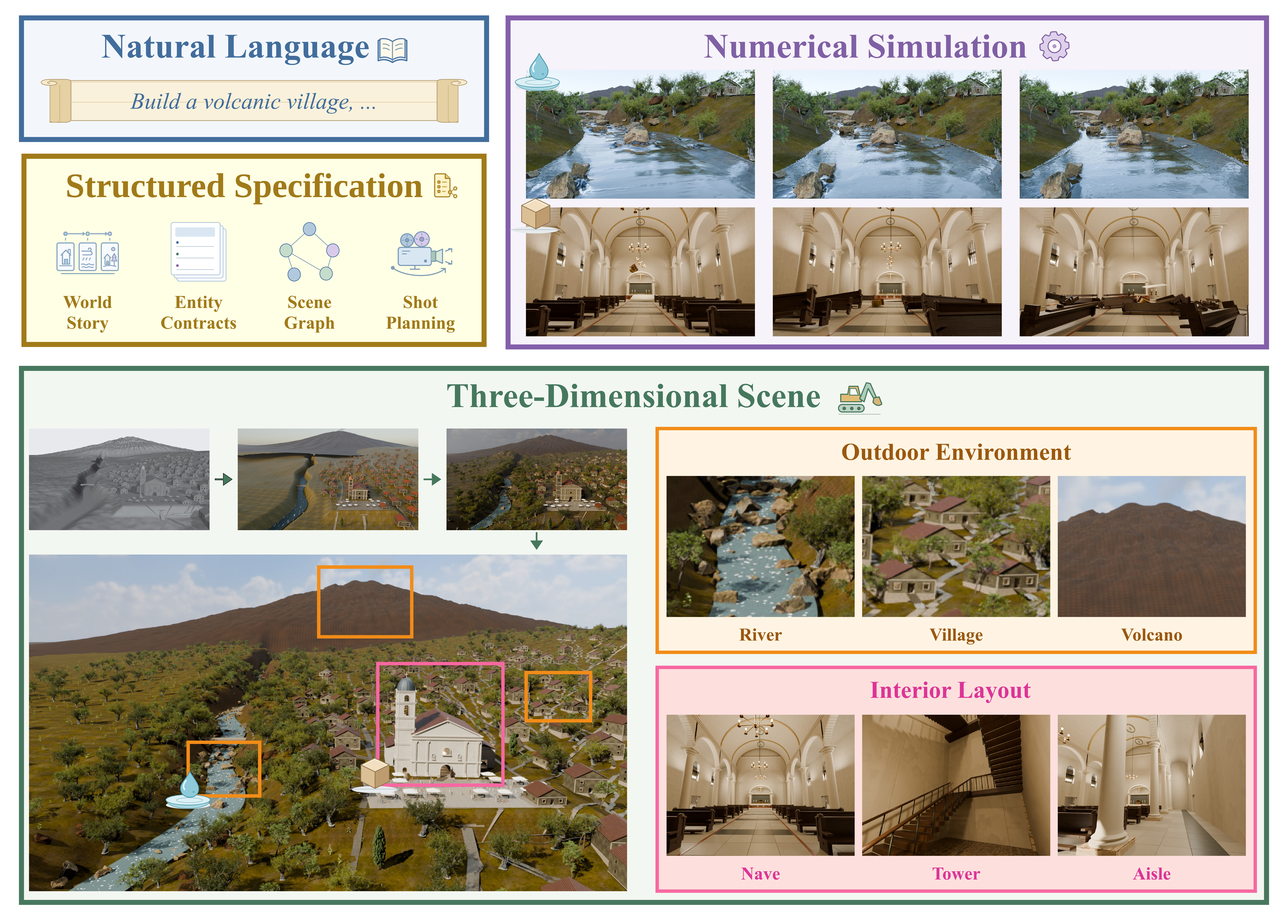}
    \caption{\textbf{Overview of WorldAgent.} WorldAgent translates natural language into a structured specification and instantiates a physically plausible world, encompassing both a three-dimensional scene and its numerical simulation.}

    \label{fig:teaser_placeholder}
\end{figure*}

\section{Related Work}
\label{sec:related-work}

\paragraph{Agentic 3D scene generation.}
3D-GPT and SceneCraft organize procedural modeling and code generation through language-driven planning~\citep{sun2023threedgpt,hu2024scenecraft}; Holodeck retrieves assets and resolves spatial layouts~\citep{yang2024holodeck}. Other systems combine scene construction with visual feedback and physical consistency checks~\citep{ling2025scenethesis,pfaff2026scenesmith,xia2026sage,wang2026physcensis}. EmbodiedGen V2 connects assets, affordances, and task environments across simulators~\citep{wang2026embodiedgen}, while WorldClaw extends agentic construction to region-scale terrain and editable open-world assets~\citep{guo2026worldclaw}. These systems advance language-driven scene composition and simulation-ready environment construction.

\paragraph{Agentic simulation and dynamic world construction.}
Agentic simulation connects simulator operation, dynamic synthesis, and execution-grounded validation. For \emph{simulator operation}, LychSim exposes scene editing, state queries, and execution to agents~\citep{ma2026lychsim}; SimWorld Studio combines engine-level coding with compiler, physics, and visual feedback to construct learning environments~\citep{kang2026simworld}. For \emph{dynamic synthesis}, PhysAgent refines force fields from point-trajectory feedback~\citep{lv2026physagent}, while GS-Agent coordinates specialized agents to configure interacting physical entities and revises their behavior using rollout feedback~\citep{zhang2026gsagent}. For \emph{execution-grounded validation}, SimWorlds and ChronoAgentic combine scene or program planning with simulator execution and state or trajectory checks~\citep{liu2026simworlds,wang2026chronoagentic}; Agentic Scientific Simulation makes assumptions and solver diagnostics explicit in an interpret--act--validate loop~\citep{lie2026agenticscientific}. These overlapping directions demonstrate how execution feedback supports simulation construction and iterative refinement. Our framework, WorldAgent, organizes scene construction and simulation around a shared structured world specification, with verification after each step guiding revisions and re-execution.

\paragraph{Physical reconstruction, assets, and visual world models.}
Complementary approaches recover geometry and physical properties from observations~\citep{pfaff2025scalablereal2sim,jiang2025phystwin}, generate editable real-to-sim variations~\citep{ranawaka2026simfoundry}, or build simulation-ready assets~\citep{cao2026physxomni}. PhysDreamer uses video priors to infer material behavior and animate 3D objects through simulation~\citep{zhang2024physdreamer}. Agentic Real2Sim refines reconstructed episodes through simulator feedback~\citep{chen2026agenticreal2sim}. Visual world models generate dynamics in pixels or latent states~\citep{bruce2024genie,valevski2024gamengen,agarwal2025cosmos}. Physical reconstruction, asset generation, and visual world modeling provide complementary resources for physical world construction, from geometry and material properties to visual synthesis.

\section{Method}
\label{sec:method}

\begin{wrapfigure}{R}{0.5\textwidth}
    \vspace{-0.5\baselineskip}
    \centering
    \includegraphics[width=\linewidth]{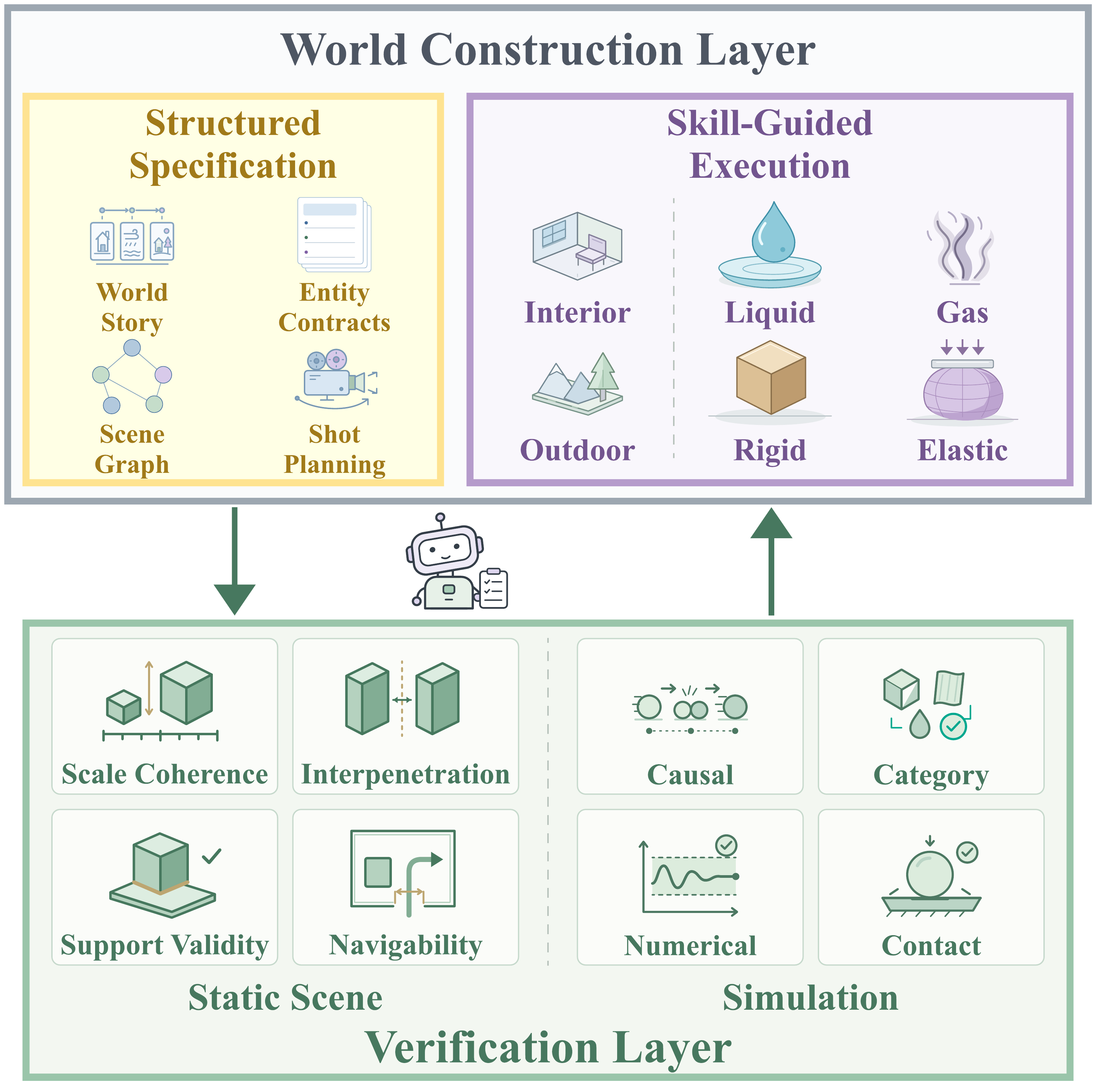}
    \caption{\textbf{Verification-guided refinement.} Green arrows indicate the feedback cycle that guides specification revisions and world construction and simulation re-execution. }
    \label{fig:verification_guided_refinement}
    \vspace{-0.5\baselineskip}
\end{wrapfigure}

From a single natural-language prompt, WorldAgent constructs a physical world comprising a 3D scene and its numerical simulation, spanning outdoor environments and interior layouts (Fig.~\ref{fig:teaser_placeholder}). For example, a prompt begins, ``Independently construct a volcanic environment at real-world scale, including a massive crater, a lava lake, \ldots'' The agent builds, simulates, verifies, and refines the world without iterative user debugging.

Our method has two layers (Fig.~\ref{fig:verification_guided_refinement}). The \emph{world construction layer} generates a structured world specification from the prompt, then follows it to build and simulate the scene in a 3D engine. After every step, the \emph{verification layer} checks geometry and simulation states alongside rendered views. Numerical simulators supply the physical behavior; the renderer supplies its views. Failed checks trigger specification revisions and re-execution of the affected steps. This self-improvement loop keeps the original user prompt fixed.

\subsection{World Construction Layer}
\label{sec:action_layer}

\subsubsection{Structured Specification}
\label{sec:proxy_representation}

Natural-language prompts leave many physical requirements implicit: a box on a table needs support, a city wall breaks at a cannon ball's hitting point, and collapses only after the hit. The agent may overlook these implicit requirements when acting directly on the prompt. Our framework therefore makes them explicit, together with the stated goals, in a structured world specification stored as a JSON manifest and organized into the following four components.


\paragraph{World story.}
The world story describes global scene semantics, action sequences, and temporal and causal dependencies. For an elastic ball that falls, rolls into a river, and produces a splash, it records the progression across entities and requires the splash to follow the ball's entry into the water.

\paragraph{Entity contracts.}
Entity contracts specify each object's role, dimensions, physical properties, and interactions. In this example, the river uses a liquid solver, the ball uses an elastic-body solver for impact deformation, and their interaction uses fluid--solid coupling. Contracts record solver inputs, outputs, and state exchange.

\paragraph{Scene graph.}
The scene graph $\mathcal{G}=(V,E)$ has objects and regions as vertices; its edges record spatial and logical relationships, including containment, support, connectivity, and interaction paths. A table supports a box; a doorway connects rooms through an opening that objects or fluid can traverse. These relations guide scene construction.

\paragraph{Shot planning.}
Shot planning specifies camera positions, viewing directions, and intended framing over time. A shot may first establish a crater and then follow lava down the volcano. Verification of framing, transitions, and event coverage guides revisions to the shot plan.

\subsubsection{Skill-Guided Execution}
\label{sec:scene_generation}

The agent follows the specification using skills developed through substantial effort informed by our simulation expertise and experimentation. Construction rules guide scene assembly; category-specific simulation knowledge guides algorithm selection, solver configuration, and failure diagnosis.

\begin{wrapfigure}{R}{0.5\textwidth}
    \vspace{-0.6\baselineskip}
    \centering
    \includegraphics[width=\linewidth]{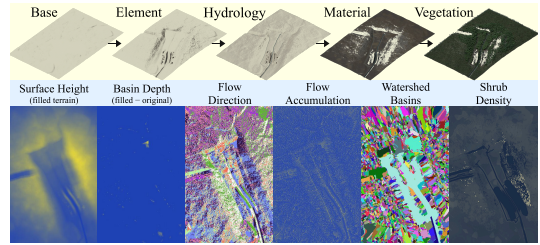}
    \caption{\textbf{Rule-guided construction for outdoor environments.} Top: terrain assembly. Bottom: hydrology-aware water and vegetation placement.}
    \label{fig:outdoor_terrain_construction}
\end{wrapfigure}

\paragraph{Rule-guided construction.}
For outdoor environments, terrain-derived fields coordinate water placement, materials, and vegetation (\cref{fig:outdoor_terrain_construction}). Priority-Flood and D8 routing identify candidate lake basins, river paths, and watersheds. Slope, sediment deposition, moisture, and exposure guide material and vegetation choices, while water and riverbed masks constrain placement. These rules establish a coordinated layout for subsequent simulation; Appendix~\ref{sec:appendix_world_reconstruction_details} gives the details.

For interior layouts, scene-graph relations become rooms, openings, and objects with specified dimensions and support. Doorways connect rooms, tables support their contents, and building entrances align with outdoor terrain. Verification of these relationships guides revisions to the layout.

\paragraph{Category-specific simulation knowledge.}
The skills cover six principal simulation categories in computer graphics---\emph{liquid}, \emph{gas}, \emph{rigid body}, \emph{elastic body}, \emph{cloth}, and \emph{particles}---plus \emph{fluid--solid coupled systems}. Setup knowledge includes PIC/FLIP blending, grid resolution, and time stepping for free-surface liquids; advection-scheme selection to preserve turbulent detail in smoke; and collision shapes and contact parameters for rigid bodies. For elastic bodies and cloth, mesh resolution balances deformation detail, computational cost, and stability, with time steps and collision distances adapted to local geometry and motion. Coupled systems specify coupling direction and state transfer between solvers.

Failure diagnosis links observed behavior to corrective actions. Unphysical water adhesion to a wall prompts inspection of boundary conditions; a rigid-body stack flying apart prompts checks for initial overlap and excessive collision margins. The agent revises the implicated scene or solver settings and tests the correction.

Short simulation probes provide feedback before expensive full runs. Once preliminary checks pass, the agent freezes the recorded settings and runs the full simulation, which also undergoes verification. Appendix~\ref{sec:supported_simulation_categories} gives implementation details.

\subsection{Verification Layer}
\label{sec:verification_layer}

The verification layer combines visual inspection of rendered results with direct examination of scene geometry and simulation states. This enables the agent to assess visible outcomes and directly check geometric and numerical properties against the structured world specification. The following eight checks cover static scene properties and simulated behavior, respectively. Relevant checks run after every construction or simulation step and over the intervals specified by the world story. Definitions and procedures appear in Appendix~\ref{sec:appendix_corefinement_details}.

\subsubsection{Static Scene Verification}

\paragraph{Scale coherence.}
The agent compares measured object dimensions with entity contracts and checks relative sizes across spatial relationships. A table and its neighboring chair, for example, should have compatible dimensions.

\paragraph{Interpenetration.}
Evaluated meshes are checked for unintended static overlap. Spatial relations and a coarse geometric search identify candidate pairs; surface queries determine penetration, with a tolerance for numerical query noise. A box embedded in a tabletop fails this check even when the camera hides the overlap.

\paragraph{Support validity.}
Required support relations must have actual contact and a continuous chain to the ground or a supporting structure. The agent checks these contacts over the intervals requiring support.

\paragraph{Navigability.}
Required scene-graph connections must form traversable routes. The agent tests the full path for sufficient width and headroom, acceptable slopes, and clearance from obstacles, including furniture blocking a doorway.

\subsubsection{Simulation Verification}

\paragraph{Causal.}
The agent checks that physical causes precede their effects as specified by the world story. For the elastic ball, water entry must precede the resulting splash. Verification examines contact, response, and their temporal order.

\paragraph{Category.}
Category-specific checks assess each physical system. Liquid verification tracks volume with prescribed inflow and outflow, particle exclusion from solids, and surface continuity. Gas verification checks density continuity and unintended loss. For elastic bodies and cloth, the agent checks deformation and displacement relative to mesh and collision scales. Coupled systems require consistent state exchange. Volume loss, for example, can reveal leakage through gaps in the ground mesh.

\paragraph{Numerical.}
The agent detects non-finite states, explosive motion, and configurations that cause instability: fluid sources inside solids, unintentionally trapped moving fluid, or overlapping solid colliders. Strong impacts with overly large time steps can severely distort deformable meshes. Diagnosis guides revisions to source placement, geometry, or solver settings, followed by a short simulation probe.

\paragraph{Contact.}
Contact verification detects unintended dynamic penetration in successive simulated states, using evaluated surfaces and the geometric criterion in Appendix~\ref{sec:appendix_corefinement_details}. A ball passing through the ground, for example, prompts revision of collision geometry or solver settings.

Failures return evidence identifying the affected requirements, objects, and frames. The agent updates the specification, re-executes affected steps, and repeats checks invalidated by the changes. Final acceptance requires reopening the same candidate and passing all required checks over the requested sequence; failed or unverified candidates remain incomplete.

\section{Experiments}
\label{sec:experiments}

\begin{figure*}[t]
    \centering
    \includegraphics[width=\textwidth]{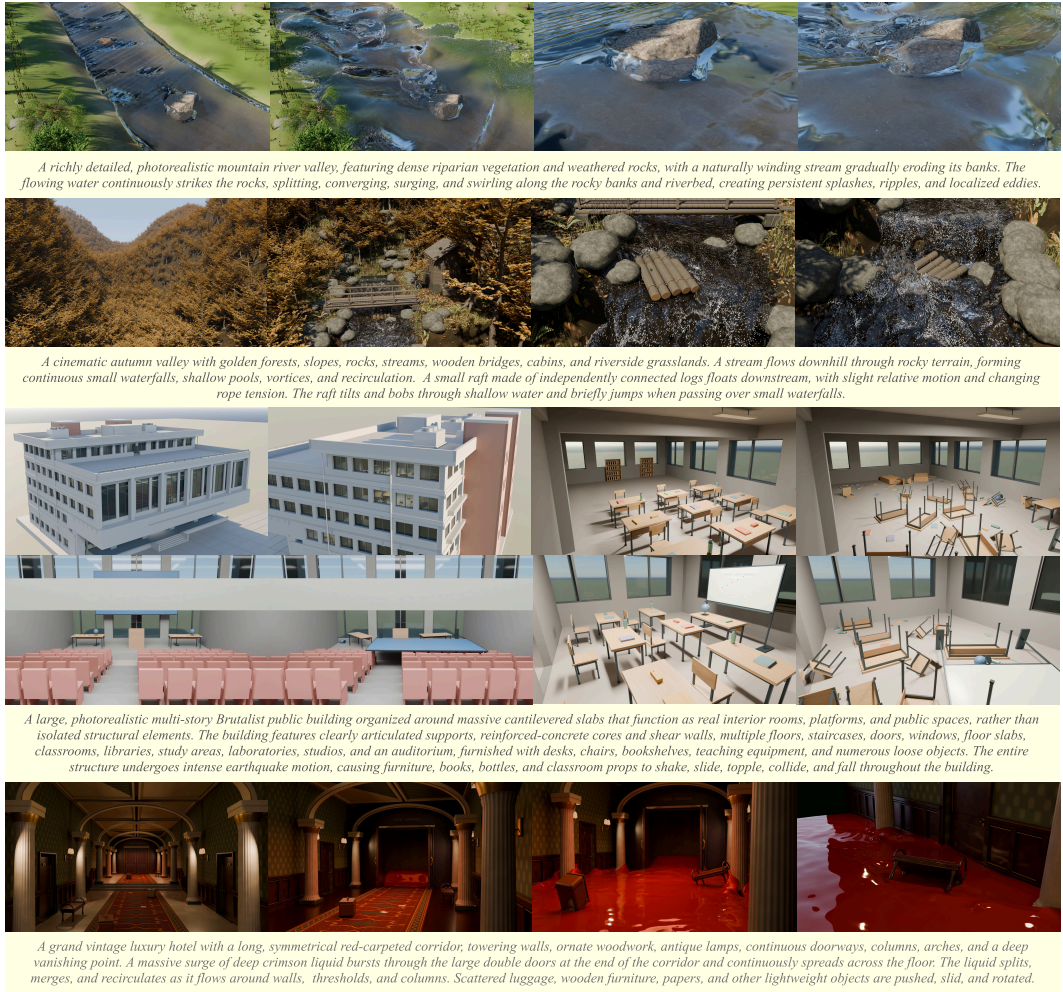}
    \caption{\textbf{WorldAgent main results.} From top to bottom, we show outdoor environments (1, 2), interior layouts (3), and cinematic scenes (4); from left to right, frames follow the temporal progression of the shot plan.}
    \label{fig:main_result_placeholder}
\end{figure*}

\subsection{Main Results}
\label{sec:main_results}

We first test our system with GPT-6 Astra as the language model backbone across a broad range of prompts. Representative sequences are shown in Figure~\ref{fig:main_result_placeholder}, covering both outdoor environments and interior layouts that exhibit rich and detailed environments, coherent spatial organization, diverse physically plausible dynamics, and strong semantic alignment with the input. We further demonstrate that our method can construct a cinematic scene from a textual description of a scenario from \emph{The Shining}. These results stem from the generality of our verification-based approach to world construction, allowing it to flexibly handle diverse environments, physical interactions, and dynamic events. Additional results are provided in Appendix~\ref{sec:appendix_additional_results}.

\subsection{Comparison with Baselines}
\label{sec:benchmark_comparison}

To further evaluate the performance of our method, we construct a new benchmark, termed \emph{AgenticSimBench}. AgenticSimBench contains 13 long-form prompts spanning outdoor environments and interior layouts with rich and compositional physical interactions. In contrast to the concise, mechanism-centered prompts used by GS-Agent and SimWorlds~\citep{zhang2026gsagent,liu2026simworlds}, which typically focus on simple physical scenarios involving few entities, each task jointly specifies a complex scene, interacting entities, and an event sequence. This design tests whether a method can construct an executable physical world in which the scene and its physical evolution jointly satisfy the instruction, rather than merely reproducing an isolated physical effect.

\begin{figure*}[t]
    \centering
    \includegraphics[width=\textwidth]{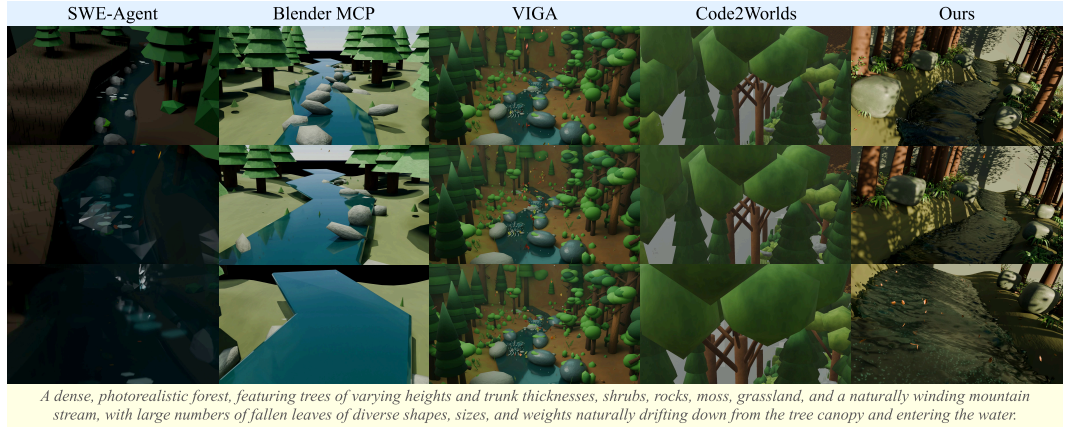}
    \caption{\textbf{Baseline comparison visualization.} All five methods receive the same prompt shown in the figure. Across all baselines, WorldAgent (Ours) produces highly detailed, spatially coherent environments with diverse, physically plausible interactions and aesthetic compositions. From top to bottom, frames follow the temporal progression of the shot plan.
}
    \label{fig:method_comparison_placeholder}
\end{figure*}

\begin{table*}[t]
    \centering
    \small
    \setlength{\tabcolsep}{4pt}
    \resizebox{\textwidth}{!}{%
    \begin{tabular}{cccccccc}
        \toprule
        & \multicolumn{2}{c}{\textbf{Physical Plausibility}} & \multicolumn{1}{c}{\textbf{Shot Planning}} & \multicolumn{2}{c}{\textbf{Content Alignment}} & \multicolumn{2}{c}{\textbf{Aesthetics}} \\
        \cmidrule(lr){2-3}\cmidrule(lr){4-4}\cmidrule(lr){5-6}\cmidrule(lr){7-8}
        Method / Metric & \mbox{Pen.  $\downarrow$} & \mbox{Caus. $\uparrow$} & \mbox{Qual. $\uparrow$} & \mbox{Comp. $\uparrow$} & \mbox{Align. $\uparrow$} & \mbox{Imag. $\uparrow$} & \mbox{Prod. $\uparrow$} \\
        \midrule
        SWE-Agent & 0.3005 & 0.2862 & 0.5436 & 0.9423 & \bestresult{16.2615} & 0.5635 & 0.7715 \\
        Blender MCP & 0.9007 & 0.3223 & 0.5487 & 0.9038 & 15.6957 & 0.6730 & 0.7103 \\
        VIGA & 0.1338 & 0.3152 & 0.4859 & 0.7500 & 16.0059 & \bestresult{0.6739} & 0.7713 \\
        Code2Worlds & 0.1148 & 0.0656 & 0.5051 & 0.5577 & 15.1408 & 0.5546 & 0.6587 \\
        Ours (GPT-5.6 Sol) & \bestresult{0.0613} & \bestresult{0.4081} & \bestresult{0.5962} & \bestresult{1.0000} & 16.1757 & 0.6707 & \bestresult{0.7866} \\
        \midrule
        Wan 2.2 & -- & 0.4692 & 0.5795 & 0.9615 & 15.2647 & 0.7070 & \bestresult{0.9310} \\
        MiniMax H3 & -- & 0.5408 & 0.5256 & 0.9615 & \bestresult{16.6525} & 0.7011 & 0.9220 \\
        Ours (GPT-6 Astra) & \bestresult{0.0415} & \bestresult{0.6825} & \bestresult{0.6042} & \bestresult{1.0000} & 16.1225 & \bestresult{0.7129} & 0.8319 \\
        \bottomrule
    \end{tabular}%
    }
   \caption{\textbf{Baseline comparison results.} Green highlights the best result for each metric in the benchmark evaluation. Top: Comparison with agent-based approaches, all using GPT-5.6 Sol as the language model backbone; our method achieves the best overall performance. Bottom: Comparison with video generation models. Video generation models lack Blender files and thus cannot be evaluated on Pen. Our method with GPT-6 Astra achieves performance comparable to video generation models. Per-metric evaluation details are provided in the Appendix~\ref{sec:appendix_benchmark}.}

    \label{tab:agent_comparison_v2}
\end{table*}

We compare our method with four other agent-based approaches. Blender MCP~\cite{ahuja2025blendermcp} represents agents that directly leverage existing tool calls, while SWE-Agent~\cite{yang2024sweagent} represents more general-purpose design-oriented agents rather than agents specifically designed for physical world construction. VIGA~\cite{yin2026viga} and Code2Worlds~\cite{zhang2026code2worlds} represent verification-based approaches that rely primarily on visual signals as feedback. All comparison experiments are conducted with GPT-5.6 Sol using the same prompt input. A visualization of one representative case is shown in Figure~\ref{fig:method_comparison_placeholder}. In our experiments, baseline outputs often assemble scenes from simple objects or use camera motion that does not clearly reveal the world's key physical behaviors. Some outputs also exhibit placement or simulation issues, such as rocks suspended without support or water represented by a geometric surface without simulated flow and interaction.

WorldAgent achieves the best scores among the compared agent-based methods on five of seven metrics using the same GPT-5.6 Sol backbone (Table~\ref{tab:agent_comparison_v2}). Relative to the strongest baseline for each metric, it reduces normalized maximum penetration by 46.6\% and improves Causal Beat Completion by 26.6\%, while achieving a Scene Composition Completeness score of 1.0. These results demonstrate its effectiveness in constructing complex physical worlds with coherent scene structure and physical behavior. Evaluation covers \emph{Physical Plausibility}, \emph{Shot Planning Quality}, \emph{Content Alignment}, and \emph{Aesthetics}, using both executable scenes and rendered videos; metric definitions appear in Appendix~\ref{sec:appendix_benchmark}.

\begin{wraptable}{R}{0.52\textwidth}
    \centering
    \small
    \setlength{\tabcolsep}{4pt}
    \resizebox{\linewidth}{!}{%
    \begin{tabular}{ccccc}
        \toprule
        \multirow{2}{*}{\centering Method} &
        Physical &
        Shot &
        Content &
        \multirow{2}{*}{\centering Aesthetics $\uparrow$} \\
        &
        Plausibility $\uparrow$ &
        Planning $\uparrow$ &
        Alignment $\uparrow$ &
        \\
        \midrule
        GPT-5.6 Sol & 2.226 & 2.811 & 2.578 & 2.293 \\
        SWE-Agent & 2.503 & 3.005 & 2.946 & 2.692 \\
        Blender MCP & 2.103 & 2.763 & 2.724 & 2.247 \\
        VIGA & 2.196 & 2.255 & 2.560 & 2.636 \\
        Code2Worlds & 2.491 & 2.241 & 2.203 & 2.369 \\
        Ours (GPT-5.6 Sol) & \bestresult{3.647} & \bestresult{3.484} & \bestresult{3.469} & \bestresult{3.615} \\
        \bottomrule
    \end{tabular}%
    }
    \caption{\textbf{User study results.} Mean ratings on a 1--5 scale across four criteria, with higher scores indicating better performance. Green highlights the highest mean.}
    \label{tab:user_study_v2}
\end{wraptable}

In a 26-participant user study, WorldAgent receives the highest mean ratings across all four criteria (Table~\ref{tab:user_study_v2}). On the five-point scale, it exceeds the highest-scoring baseline in each criterion by 1.14 points in Physical Plausibility, 0.48 in Shot Planning Quality, 0.52 in Content Alignment, and 0.92 in Aesthetics. Participants compare videos from all six methods on randomly assigned AgenticSimBench cases and rate each criterion on a 1--5 scale. The study protocol is detailed in Appendix~\ref{sec:appendix_user_study_details}.


To further evaluate the visual quality of the physical worlds generated by our system, we compare our rendered outputs with MiniMax H3 and Wan 2.2~\cite{wan2025} on the same 13 prompts from AgenticSimBench. As shown in the lower block of Table~\ref{tab:agent_comparison_v2}, our method achieves visual quality comparable to state-of-the-art open-source video generation models. More importantly, our generated worlds are not only visually compelling, but also controllable, editable, and physically grounded.

\subsection{Ablation Studies}
\label{sec:ablation_v2}

\begin{figure*}[t]
    \centering
    \includegraphics[width=\textwidth]{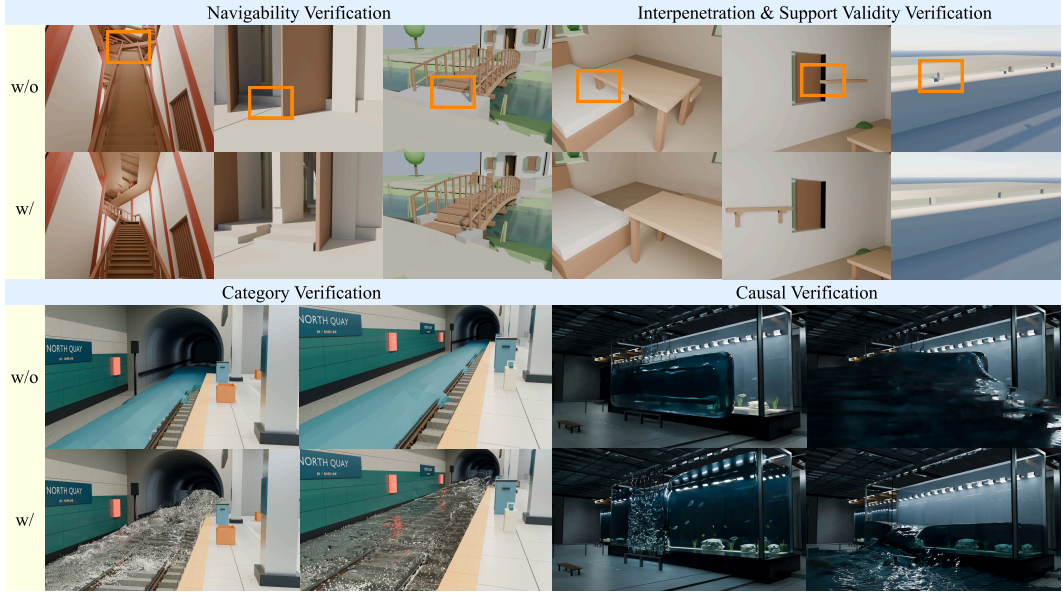}
    \caption{\textbf{Ablation of verification.} Orange boxes highlight failed details. Top left: w/o, malformed stairs, blocked exits caused by oversized thresholds, and abrupt transitions at required connections; w/, all routes remain traversable. Top right: w/o, the bed and table interpenetrate, the window and platform interpenetrate, and the balcony pipe lacks support; w/, no interpenetration or missing support. Bottom left: w/o, coarse, patch-like blue fluids exhibit unnatural motion; w/, the fluids appear realistic with smooth motion. Bottom right: w/o, glass breaking and aquarium leakage occur simultaneously without causal consistency; w/, glass breaking causes the subsequent water leakage.}

    \label{fig:ablation_verification}
\end{figure*}

Figure~\ref{fig:ablation_verification} shows the effects of removing verification. Removing navigability checks produces blocked or discontinuous routes; removing interpenetration and support checks produces overlapping or unsupported objects. Category verification guides the simulation toward correct and aesthetically coherent realizations of the intended physical behaviors. Removing it leads to coarse, patch-like blue fluids with unnatural motion, whereas category verification produces realistic fluids with smooth motion. Causal verification checks the dependencies recorded in the world story: without it, glass breaking and aquarium leakage may occur simultaneously instead of in the required causal order.

\begin{table*}[t]
    \centering
    \small
    \setlength{\tabcolsep}{4pt}
    \resizebox{\textwidth}{!}{%
    \begin{tabular}{lccccccc}
        \toprule
        & \multicolumn{2}{c}{\textbf{Physical Plausibility}} & \multicolumn{1}{c}{\textbf{Shot Planning}} & \multicolumn{2}{c}{\textbf{Content Alignment}} & \multicolumn{2}{c}{\textbf{Aesthetics}} \\
        \cmidrule(lr){2-3}\cmidrule(lr){4-4}\cmidrule(lr){5-6}\cmidrule(lr){7-8}
        \multicolumn{1}{c}{Method / Metric} & \mbox{Pen.  $\downarrow$} & \mbox{Caus. $\uparrow$} & \mbox{Qual. $\uparrow$} & \mbox{Comp. $\uparrow$} & \mbox{Align. $\uparrow$} & \mbox{Imag. $\uparrow$} & \mbox{Prod. $\uparrow$} \\
        \midrule
        GPT-5.6 Sol & 0.6747 & 0.3004 & 0.5038 & 0.7692 & 15.6923 & 0.5836 & 0.6419 \\
        GPT-6 Astra & 0.2248 & 0.6031 & 0.5308 & 0.9615 & 16.2150 & \bestresult{0.7296} & 0.8465 \\
        Opus 5 & 0.8316 & 0.4255 & 0.5300 & 0.9500 & 17.4162 & 0.6368 & 0.8226 \\
        \midrule
        Ours (GPT-5.6 Sol) & 0.0613 & 0.4081 & 0.5962 & \bestresult{1.0000} & 16.1757 & 0.6707 & 0.7866 \\
        Ours (GPT-6 Astra) & \bestresult{0.0415} & \bestresult{0.6825} & \bestresult{0.6042} & \bestresult{1.0000} & 16.1225 & 0.7129 & 0.8319 \\
        Ours (Opus 5) & 0.2038 & 0.6638 & 0.6000 & \bestresult{1.0000} & \bestresult{17.6531} & 0.6505 & \bestresult{0.8489} \\   
        \bottomrule
    \end{tabular}%
    }
    \caption{\textbf{Ablation of verification and language model backbone.} Green highlights the best result for each metric in the benchmark evaluation. Rows named only by the backbone directly feed the same prompts to the corresponding agent, whereas Ours applies the complete WorldAgent framework with that backbone. Across the evaluated backbones, WorldAgent consistently improves over the corresponding backbone-only agent, demonstrating the effectiveness of our framework across language model backbones.}

    \label{tab:ablation_v2}
\end{table*}

We also conduct ablation studies across different language model backbones by comparing each backbone-only agent with WorldAgent using the same backbone and the metrics from Sec.~\ref{sec:benchmark_comparison}. Table~\ref{tab:ablation_v2} reports complete comparisons for GPT-5.6 Sol, GPT-6 Astra, and Opus 5. Additional qualitative comparisons between methods are provided in Appendix~\ref{sec:appendix_additional_results}.

\section{Conclusion}
\label{sec:conclusion}

We presented WorldAgent, a framework for verification-guided physical world construction organized into a world construction layer and a verification layer. AI agents use a structured world specification, including entity contracts, alongside category-specific simulation knowledge and access to engine state to build scenes and run simulations. Verification after each step identifies failures and guides revisions to the specification, followed by re-execution of the affected steps. Simulation thus produces both the world's physical behavior and evidence for refining its scene and solver settings.

The current system remains bounded by Blender's solvers, imperfect physical-parameter inference, and the cost of strict full-trajectory verification. Some granular, fracture, combustion, and multi-physics effects therefore require disclosed approximations or one-way coupling. Extending the structured world specification and verification process across specialized backends, and learning from both accepted and failed simulations, are direct paths toward AI agents that construct editable, physically testable worlds.

\subsection*{AI Use Statement}

AI tools were used to improve the clarity and readability of the manuscript and to assist with literature searches. We explore the capabilities of AI agents  on a set of predefined physical-world construction scenarios designed by the authors. The authors take full responsibility for the final work.

\subsection*{Reproducibility Statement}

We will release our agent skills. Methodological and evaluation details are provided in the paper and appendix.

\bibliography{main}
\bibliographystyle{iclr2027_conference}

\appendix
\clearpage
\providecommand{\comparisonbreak}{}
\section{Method Details}
\label{sec:appendix_method_details}

The world construction layer builds and simulates the physical world from a structured world specification. Verification follows every step; failed checks trigger specification revisions and re-execution of affected steps, with the user prompt fixed throughout.

\subsection{World Construction Layer}

\subsubsection{Structured Specification}
\label{sec:appendix_proxy_details}

We use the hotel-corridor flooding scene in Figure~\ref{fig:main_result_placeholder}(4) to illustrate how the structured world specification connects requirements to execution and verification.

\paragraph{World story.} This excerpt from world story in the manifest file below associates a prompted scene with identifiers, responsible solvers or inputs, frame intervals, and acceptance predicates. It describes liquid descending across a stepped landing:
{\small
\begin{verbatim}
{
  "id": "b4_cascade",
  "owner": "Mantaflow liquid + three-step landing",
  "frames": [80, 150],
  "predicate":
    "liquid crosses the 3 x 0.15 m steps and falls to the lower hall"
}
\end{verbatim}
}
Here, \texttt{frames} specifies the inspection interval and \texttt{predicate} the required outcome: liquid crossing the landing during that interval.

\paragraph{Entity contracts.} In the same case, our manifest file assigns liquid motion to Mantaflow liquid configured with the FLIP method, and furniture motion to rigid-body simulation. The world construction layer also authored inflow switches and forces. These assignments link each object to its physical parameters and the geometry used for rendering and collision detection, guiding inspection after a failed check.

\paragraph{Scene graph.} The manifest names the upper hall, stepped landing, and lower hall. The stepped landing connects the floor levels, while recorded support relations link objects to their supporting surfaces.

\paragraph{Shot planning.} The shot plan specifies camera viewpoints, framing, and transitions over time, keeping key interactions and outcomes visible during the relevant story intervals.

\paragraph{Specification audit.} Before expensive execution, the agent checks that referenced entities and connection endpoints exist, motion has declared solver ownership, and each mandatory requirement has a checkable predicate and evidence source. These records guide construction, solver dependencies and bake order, contact-pair selection, and camera coverage. Missing or unresolved requirements remain \emph{unverified}.

\subsubsection{Skill-Guided Execution}
\label{sec:appendix_world_reconstruction_details}

\paragraph{Rule-guided construction.}
The agent expands the scene graph from landmarks or functional zones with metric bounds and connectivity to rooms, paths, openings, supports, and objects. Compact scenes use fewer levels.

Outdoor construction coordinates water, materials, and vegetation through terrain-derived fields (\cref{fig:outdoor_terrain_construction}). Priority-Flood fills depressions in heightmap $H$ to obtain $H_f$ (\emph{Surface Height}), leaving terrain geometry unchanged. Connected components of $H_f-H$ (\emph{Basin Depth}) identify lake basins, characterized by depth, storage volume, and contributing area. D8 routing on $H_f$ yields \emph{Flow Direction} and \emph{Flow Accumulation}; upstream tracing identifies \emph{Watershed Basins} and drainage divides. The requested biome and water-system density guide basin selection and channel thresholds. Rivers follow high-accumulation paths; streams descend from candidate sources near divides to reachable rivers or lakes.

Material rules combine these fields with slope, elevation, curvature, gradient-based roughness, local maxima, estimated solar exposure, and erosion and deposition diagnostics. At each cell, the last matching rule determines the material, with a default for unmatched cells: steep slopes can receive exposed rock and depositional channels sediment. Moisture estimates combine water proximity, material water retention, flow accumulation, and prescribed rainfall. Vegetation rules combine moisture with bank distance, accumulated sunlight, terrain shelter, and wind exposure to set types and density. Water and riverbed masks enforce placement exclusions; \emph{Shrub Density} in \cref{fig:outdoor_terrain_construction} shows the resulting distribution. Interior layouts translate connections into openings with appropriate clearance and supports; building entrances align with outdoor ground levels.

Procedural asset generation proceeds from primary form to secondary structure and camera-visible detail. Refinement preserves identifiers, placement, support relations, and scene-graph roles. Correspondences among meshes used for rendering, simulation, and collision detection preserve relevant contact features.

\paragraph{Category-specific simulation knowledge.}
\label{sec:supported_simulation_categories}
The skills encode algorithm selection, parameter configuration, failure diagnosis, and corrective action. Records identify Blender objects and collections, solver inputs and outputs, collision representations, and cache ranges.

For \emph{liquid}, we use FLIP-based simulation through either the FLIP Fluids add-on or Mantaflow liquid, configuring the PIC/FLIP ratio, grid resolution, and time step for the selected implementation. Grids resolve sources, obstacles, and narrow passages. Low-resolution probes establish timing; production-resolution output establishes final surface quality. For \emph{gas}, Mantaflow supports smoke and fire, with advection-scheme selection to retain turbulent detail and appropriate source, buoyancy, and vorticity settings.

For \emph{rigid bodies}, Bullet uses specified mass, collision shapes and margins, friction, constraints, and temporal resolution. Fracture requires a breakable representation. For \emph{elastic bodies}, Soft Body provides spring-based deformation, with recorded rest topology, spring or binding structure, and deformation and recovery bounds. Its accuracy is limited to this spring-based approximation. For \emph{cloth}, settings include mesh resolution, pins, stiffness, damping, thickness, collider layers, and self-collision where needed. Elastic-body and cloth settings balance resolution, cost, and stability, adapting time steps and collision distances to thickness, local edge lengths, and motion per step.

For \emph{particles}, the implementation is limited to ballistic or granular approximations, recording particle count, occupied volume, velocity, and settling behavior. For \emph{fluid--solid coupled systems}, the specification declares coupling direction, exchanged variables, state ownership, driving boundaries, and bake order. Two-way coupling requires exchanged reaction forces; one-way approximations and omitted reactions are reported explicitly.

Failure diagnosis follows the simulation evidence. Unphysical liquid adhesion to a wall prompts boundary-condition correction. For smoke, density and domain evidence distinguish solver loss from material fading or camera cropping. A rigid-body stack that flies apart prompts checks for initial overlap and excessive collision margins. The agent revises the implicated scene or solver settings and tests the correction.

Short probes cover preroll, activation, first contact, peak response, and early settling. After a probe passes verification, its configuration is frozen and baked to an isolated candidate cache in the declared solver order. Acceptance then requires verification of the full simulated sequence.

\subsection{Verification Layer}
\label{sec:appendix_corefinement_details}

After every construction or simulation step, a requirements matrix records each predicate, status, inspected artifact and frame interval, measured evidence, first failing object or frame, and corrective action. Rows cover the specification's spatial, physical, temporal, observational, and delivery requirements. Progression requires evidence-backed passes for all mandatory rows.

\subsubsection{Static Scene Verification}

\paragraph{Scale coherence.} Scalar characteristic sizes $d_i$ and $d_j$ are derived from the specified dimensions of objects $i$ and $j$ using the same measure and physical units, such as each bounding box's diagonal length. For a spatial adjacency or contact relation $(i,j)$, we require
\begin{equation}
    \rho_{ij} = \frac{d_i}{d_j}
    \in [\rho_{\min}(i,j), \rho_{\max}(i,j)],
    \label{eq:scene_scale_coherence}
\end{equation}
with bounds determined by entity contracts and spatial relationships. Measurements of constructed geometry also verify the specified dimensions.

\paragraph{Interpenetration.} Spatial relationships and bounding boxes identify potentially interacting pairs; evaluated surfaces determine unintended static overlap. Let $\mathrm{dist}(i,j)$ denote signed mesh clearance, negative for penetration. The check requires
\begin{equation}
    \mathrm{dist}(i,j) \ge -\epsilon_{\mathrm{pen}},
    \label{eq:scene_interpenetration}
\end{equation}
where $\epsilon_{\mathrm{pen}}$ accounts solely for numerical noise in the geometry query. Detected static penetration fails verification.

\paragraph{Support validity.} Verification checks actual contact and continuous support chains over the intervals requiring support.

\paragraph{Navigability.} Verification checks the width, headroom, slope, and continuity of required routes.

\subsubsection{Simulation Verification}

\paragraph{Causal.} Contact and response must follow the world story's temporal dependencies: the elastic ball enters water before causing a splash. Verification checks both the order and the causal connection. Required triggers, activation, responses, and outcomes must remain observable from the specified cameras.

\paragraph{Category.} Liquid checks track volume with prescribed inflow and outflow, keep particles outside solid interiors, and detect leakage through unresolved gaps. Surface checks identify spurious chunks, spikes, discontinuous components, and frame-to-frame popping. Gas checks test density continuity and unintended loss or escape. Elastic-body and cloth checks measure deformation and displacement relative to local mesh and collision scales, including self-intersection and failed recovery. Coupled systems require consistent transferred states and recorded coupling direction. These checks use the corresponding category-specific records.

\paragraph{Numerical.} Probes and full sequences are checked for non-finite coordinates, explosive growth, extreme displacement or velocity, and invalid active counts. Diagnosis covers fluid sources inside solids, unintentionally trapped moving fluid, overlapping rigid colliders, and excessive deformation within one time step. The agent revises temporal resolution, collision parameters, or responsible source placement and geometry, then reruns a short probe.

\paragraph{Contact.} Contact verification checks for unintended dynamic interpenetration during simulation. At each frame, the agent applies the same geometric criterion as static interpenetration (Eq.~\ref{eq:scene_interpenetration}) to potentially contacting objects and the surrounding scene. Detected penetration prompts revisions to collision geometry or solver settings.

\paragraph{Feedback and acceptance.} Failed checks return to the world construction layer for specification revisions. Geometry, scale, support, openings, and collision-shape failures require scene changes; temporal resolution, material response, source inputs, and coupling failures require solver changes. Each iteration records a hypothesis and changes one causal parameter group before a short probe. Physical failures are repaired through the implicated scene or solver configuration. Changes to participating geometry, solvers, driving inputs, coupling, or simulation intervals invalidate dependent caches and evidence; unaffected components are retained.

For final acceptance, the exact candidate is reopened in a fresh Blender process for read-only verification over the complete requested frame range. It must pass all eight checks and meet the shot-planning requirements, with geometry and simulator-state evidence retained. Failed or unverified candidates remain incomplete and are preserved with useful diagnostics. A passing candidate proceeds to final rendering.

\section{Experiment Prompt}
\label{sec:appendix_prompts}

All evaluated agent baselines receive and must follow the same prompt below. Only the knowledge-file path, number of scenes, scene descriptions, and independent output-folder names are instantiated for each run.

\begin{quote}
\small
Strictly follow the knowledge, specifications, available tool capabilities, and related instructions provided in \texttt{<PATH\_TO\_SKILL.md>}, and iteratively complete the following physical-simulation scenes.

\noindent\textbf{Important requirements:}
\begin{itemize}
    \item Do not inspect, read, reference, or reuse any existing scene, script, output, result, or other pre-existing file in that directory. Use only the knowledge provided in the directory as the basis for scene production.
    \item The scenes must be completely independent. Build each scene from scratch with its own environment, objects, materials, physical state, simulation process, caches, and outputs.
    \item No scene may depend on, reference, inherit, or reuse any model, object, material, animation, physics cache, simulation result, or intermediate file from another scene.
    \item Each scene must open, run, and render independently, without loading, executing, or depending on any other scene.
    \item Even when scenes contain similar object types or physical phenomena, create and simulate those elements independently within each scene.
    \item Make every scene conform as closely as possible to real-world physics and pursue cinematic realism through plausible scale, materials, lighting, motion, collision, deformation, and environmental detail.
    \item Do not over-simplify a scene merely to reduce computation. Within the available capability, increase object count, geometric detail, and the complexity of physical interactions.
    \item Place all newly generated scenes, scripts, caches, and rendered outputs in newly created, scene-specific subfolders under \texttt{<PATH\_TO\_OUTPUT>}. Do not overwrite or modify any existing file.
    \item Every file in a subfolder may serve only its corresponding scene and must not be shared across scenes.
    \item Give the scenes clearly different visual content, environments, and physical processes; do not repeat the same simulation setup with superficial changes.
    \item Give each scene a clear cinematic shot design, and derive the visual result from real physical processes rather than relying only on preset animation or simple visual effects.
    \item Whenever possible, use Blender's actual Rigid Body, Rigid Body Constraint, Cloth, Soft Body, Fluid/Mantaflow, Smoke, Fire, Particle System, Force Field, Collision, Dynamic Paint, Geometry Nodes, Ocean, Wave, and fracture or pre-fractured geometry capabilities.
    \item For large-scale destruction, prioritize real fractured geometry, pre-segmented structures, rigid-body constraints, and collision rather than hiding or moving an intact model with simple keyframes.
    \item For fluids, smoke, fire, cloth, deformable objects, and particles, use the corresponding physical simulation systems available in Blender whenever possible.
    \item In complex multi-physics scenes, establish genuine causal relationships between simulations rather than stacking independent visual effects.
    \item Use plausible real-world scale, gravity, mass, friction, restitution, damping, collision thickness, and simulation time steps in every scene.
    \item Give each scene its own cameras, lights, materials, simulation caches, and outputs.
    \item After completing a scene, confirm that it can be opened, recomputed, and rendered independently before proceeding to the next scene.
\end{itemize}

\noindent\textbf{Scene prompts:}

\noindent[Prompts for \texttt{example\_1}, \texttt{example\_2}, $\ldots$, \texttt{example\_N}]

Finally, save the $N$ scenes separately under \texttt{<PATH\_TO\_OUTPUT>} in $N$ newly created and completely independent subfolders:

\begin{verbatim}
outputs/
|-- example_1/
|-- example_2/
|-- ...
`-- example_N/
\end{verbatim}

For every scene, continue improving asset realism, material realism, rendering realism, story completeness, cinematic narration, smooth shot transitions, scene richness, and overall finish until the result reaches a satisfactory and evidence-supported level. Do not use external 3D models, although scripts may generate non-template, high-fidelity meshes. Render a complete 1920x1080 final sequence with Cycles and OptiX. Complete and validate the current example before proceeding to the next example in batch order, and continue until all examples have been completed, validated, and rendered as final sequences.
\end{quote}

\section{Benchmark Comparison Details}
\label{sec:appendix_benchmark}

We evaluate all methods on the same $N=13$ task prompts. For method $m$ and task $i$, let $V_{m,i}$ denote the rendered video, $P_i$ the prompt, and $B_{m,i}$ the executable Blend artifact when available. Unless stated otherwise, the reported method score is the task-level macro average
\begin{equation}
    S_m = \frac{1}{N}\sum_{i=1}^{N}s(V_{m,i},P_i).
    \label{eq:metric_macro_average}
\end{equation}
Video-only methods receive all render-based scores; geometry-based penetration is recorded as unavailable. VLM rubrics and auxiliary questions are fixed across methods before aggregation.

\subsection{Normalized Maximum Penetration Depth}
\label{sec:metric_penetration}

This geometry-based metric measures the worst dynamic penetration over sampled simulated states in an executable scene. We uniformly sample $K_p=24$ frames over the complete animation interval of $B_{m,i}$ and evaluate the dependency-graph mesh at each sampled frame, thereby including animation, modifiers, and simulator output. Axis-aligned bounding boxes provide only a broad-phase candidate set. For every candidate dynamic-object--collider pair $(a,b)$, Blender's bounding-volume hierarchy is used to require an actual triangle--triangle overlap before a penetration depth is recorded.

For a vertex $v$ incident to an overlapping triangle of $a$, let $\pi_b(v)$ and $n_b(v)$ be its nearest point and consistently oriented surface normal on $b$. The directed depth is
\begin{equation}
    d_{a\rightarrow b}
    = \max_{v\in\mathcal{V}_{a\cap b}}
      \left[ -\bigl(v-\pi_b(v)\bigr)^\top n_b(v) \right]_+,
    \label{eq:directed_penetration_depth}
\end{equation}
where $[x]_+=\max(x,0)$. We evaluate both directions when both objects are dynamic, sample at most 128 incident vertices per direction, and use $10^{-4}$\,m as the minimum positive depth for a detected triangle overlap. Let $L_{m,i}$ be the diagonal of the evaluated scene bounds and $\mathcal{C}_{m,i,t}$ the overlapping candidate pairs at frame $t$. The per-task normalized maximum penetration is
\begin{equation}
    s_{\mathrm{Pen}}(B_{m,i})
    = \frac{100}{L_{m,i}}
      \max_{t\in\mathcal{F}_i}
      \max_{(a,b)\in\mathcal{C}_{m,i,t}}
      \max\!\left(d_{a\rightarrow b},d_{b\rightarrow a}\right).
    \label{eq:normalized_max_penetration}
\end{equation}

Normalized Maximum Penetration Depth measures the severity of object interpenetration in the constructed world. Lower is better.

\subsection{Causal Beat Completion}
\label{sec:metric_causal_completion}

We use an anonymized LLaVA-Video-7B-Qwen2 reviewer~\citep{zhang2024videoinstruction} to assess whether the prompt-critical physical event chain is visibly completed. Each task has five manually specified causal beats. We divide every video into $S_c=5$ ordered narrative stages, sampling at most $K_c=16$ frames per stage with a 10\% temporal overlap at stage boundaries. For stage $s$, the VLM reviewer assigns $b_{m,i,s}\in\{0,1/2,1\}$ for the corresponding beat: absent, partial or unclear, or clearly completed. For each adjacent stage pair, the reviewer assigns $\ell_{m,i,s}\in\{0,1/2,1\}$ for the causal connection: no or wrong-order connection, plausible order but unclear linkage, or a clear response/consequence. The per-video score combines event completion and adjacent-stage causal linkage:
\begin{equation}
\begin{aligned}
e_{m,i}
  &= \frac{1}{S_c}\sum_{s=1}^{S_c} b_{m,i,s},
&\qquad
\ell_{m,i}
  &= \frac{1}{S_c-1}\sum_{s=1}^{S_c-1}\ell_{m,i,s},\\
c_{m,i}
  &= 0.7e_{m,i}+0.3\ell_{m,i},
&\qquad
S_{\mathrm{Causal}}(m)
  &= \frac{1}{N}\sum_{i=1}^{N}c_{m,i}.
\end{aligned}
\label{eq:causal_completion}
\end{equation}
The same fixed rubric is applied independently to each method video, without using method names in the prompt. This score measures visible narrative completion and causal evidence. Higher is better.

\subsection{Shot Planning Quality}
\label{sec:metric_shot_planning_quality}

We divide every rendered video into $S_c=8$ contiguous temporal segments and sample at most $K_c=16$ frames within each segment. Each segment is provided separately to LLaVA-Video-7B-Qwen2~\citep{zhang2024videoinstruction}, preserving temporal coverage for long videos while allowing short transitions and speed changes to be inspected. A fixed rubric asks for three independent integer scores in $\{1,2,3,4,5\}$: $r_{m,i,s}$ for purposeful richness and variation of planned viewpoints, $t_{m,i,s}$ for naturalness and continuity of shot transitions, and $a_{m,i,s}$ for smoothness and plausibility of camera speed and acceleration. The reviewer is instructed to judge shot planning only, ignoring object, fluid, and particle motion. Each subscore is normalized to $[0,1]$, and the per-video shot-planning-quality score is
\begin{equation}
    s_{\mathrm{ShotQ}}(V_{m,i})
    = \frac{1}{S_c}\sum_{s=1}^{S_c}
      \frac{1}{3}\left(\frac{r_{m,i,s}}{5}+\frac{t_{m,i,s}}{5}+\frac{a_{m,i,s}}{5}\right).
    \label{eq:shot_planning_quality}
\end{equation}
The reported method score is the macro average of $s_{\mathrm{ShotQ}}$ over its videos. The rubric rewards purposeful viewpoint selection, coherent shot transitions, and smooth starts, stops, and speed changes. Higher is better.

\subsection{Scene Composition Completeness}
\label{sec:metric_composition}

We adapt the Composition evaluator from VBench-2.0~\citep{zheng2025vbench2}. For each of the 13 fixed prompts, we manually convert the required entities and scene regions into a list $\mathcal{Q}^{\mathrm{comp}}_i$ of binary presence questions. We divide each video into $S_c=8$ temporal segments and sample one frame every $\Delta_c=10$ source frames within each segment. If a segment contains more than $K_c=64$ sampled frames, we evaluate overlapping windows of at most $K_c$ sampled frames with a stride of 32 sampled frames. LLaVA-Video-7B-Qwen2~\citep{zhang2024videoinstruction} answers each question independently for each temporal window. If $y_{m,i,q,w}\in\{0,1\}$ denotes a positive answer, an entity counts as visible when it is detected in at least one window, $\bar y_{m,i,q}=\max_w y_{m,i,q,w}$, and the per-video score is
\begin{equation}
    s_{\mathrm{Comp}}(V_{m,i})
    = \frac{1}{|\mathcal{Q}^{\mathrm{comp}}_i|}
      \sum_{q\in\mathcal{Q}^{\mathrm{comp}}_i} \bar y_{m,i,q}.
    \label{eq:composition_completeness}
\end{equation}
We then apply Eq.~\ref{eq:metric_macro_average}. The score measures whether requested components are visibly present somewhere in the video. Higher is better. 

\subsection{Video--Text Alignment}
\label{sec:metric_alignment}

Following the Alignment Score used by GS-Agent~\citep{zhang2026gsagent}, we encode every decoded video frame and the task text with the Meta Perception Encoder~\citep{bolya2025perception}. Our fixed implementation uses the \texttt{PE-Core-L14-336} checkpoint. Frames are processed in batches for computational efficiency, but no temporal subsampling is applied. Let $T_{m,i}$ be the number of decoded frames, and let $\hat f_{m,i,t}$ and $\hat g_i$ be the L2-normalized frame and text embeddings. We compute
\begin{equation}
    s_{\mathrm{Align}}(V_{m,i},P_i)
    = \frac{100}{T_{m,i}}\sum_{t=1}^{T_{m,i}}
      \hat f_{m,i,t}^{\top}\hat g_i.
    \label{eq:video_text_alignment}
\end{equation}
Higher values indicate stronger frame-level semantic agreement with the task description.

\subsection{VBench Imaging Quality}
\label{sec:metric_imaging_quality}

We follow the original VBench Imaging Quality protocol~\citep{huang2024vbench}, using the MUSIQ multi-scale image-quality transformer~\citep{ke2021musiq} with the SPAQ checkpoint. Every decoded video frame contributes to the score. Frames whose longer side exceeds 512 pixels are resized isotropically to a maximum side length of 512, while smaller frames retain their native resolution. If $q_{\mathrm{MUSIQ}}(I_t)$ is the predicted image-quality score for frame $I_t$, we report
\begin{equation}
    s_{\mathrm{Imaging}}(V_{m,i})
    = \frac{1}{100T_{m,i}}
      \sum_{t=1}^{T_{m,i}}q_{\mathrm{MUSIQ}}(I_t),
    \label{eq:vbench_imaging_quality}
\end{equation}
followed by the task macro average. Imaging Quality measures perceptual sharpness, noise, and visible distortions. Higher is better. 

\subsection{VLM Production Quality}
\label{sec:metric_production_quality}

The production-quality score uses an anonymized LLaVA-Video-7B-Qwen2 reviewer~\citep{zhang2024videoinstruction} over the same task set. To include every decoded frame without exceeding the VLM context, we partition each video into $W_{m,i}=\lceil T_{m,i}/64\rceil$ contiguous, non-overlapping windows. Every frame appears in exactly one window, and the final window retains all remaining frames. For each window $w$, the reviewer receives the frames in temporal order and the same case-specific benchmark text, but not the method identity. It assigns four integer scores $r_{m,i,w,c}\in\{1,2,3,4,5\}$ for fidelity to the visible requested content, composition and readability, temporal identity and artifact consistency, and overall production quality. We first compute the window score
\begin{equation}
    p_{m,i,w}=\frac{1}{20}\sum_{c=1}^{4}r_{m,i,w,c}.
    \label{eq:production_quality_window}
\end{equation}
Let $n_{m,i,w}$ denote the number of frames in window $w$. The per-video score weights each window by its actual frame count,
\begin{equation}
    p_{m,i}=\frac{1}{T_{m,i}}
    \sum_{w=1}^{W_{m,i}}n_{m,i,w}p_{m,i,w},
    \qquad
    S_{\mathrm{Prod}}(m)=\frac{1}{N}\sum_{i=1}^{N}p_{m,i}.
    \label{eq:production_quality}
\end{equation}
This study-specific VLM diagnostic measures overall presentation quality. Higher is better. 

\section{User Study Details}
\label{sec:appendix_user_study_details}

Participants consented to the questionnaire study; no personal information was retained.
The study includes 26 participants and rendered outputs from GPT-5.6 Sol with direct prompting, Blender MCP, our method, SWE-Agent, Code2Worlds, and VIGA. Each participant sees eight randomly selected AgenticSimBench tasks, with all six methods' outputs and the initial prompt for each task. Videos share a presentation format and omit method names. Method and task order are randomized independently.
Participants rate each video on a five-point integer scale (1--5, higher is better) for Physical Plausibility, Shot Planning Quality, Content Alignment, and Aesthetics. These criteria cover motion, contact, deformation, and material response; purposeful viewpoint variety, transition naturalness, and speed/acceleration smoothness; scene and simulation conformity to the prompt; and overall visual quality, respectively. 
For each method and criterion, we first average each participant's ratings and then average these participant-level means. 

\section{Additional Results}
\label{sec:appendix_additional_results}

\begin{figure*}[t]
    \centering
    \includegraphics[width=\textwidth]{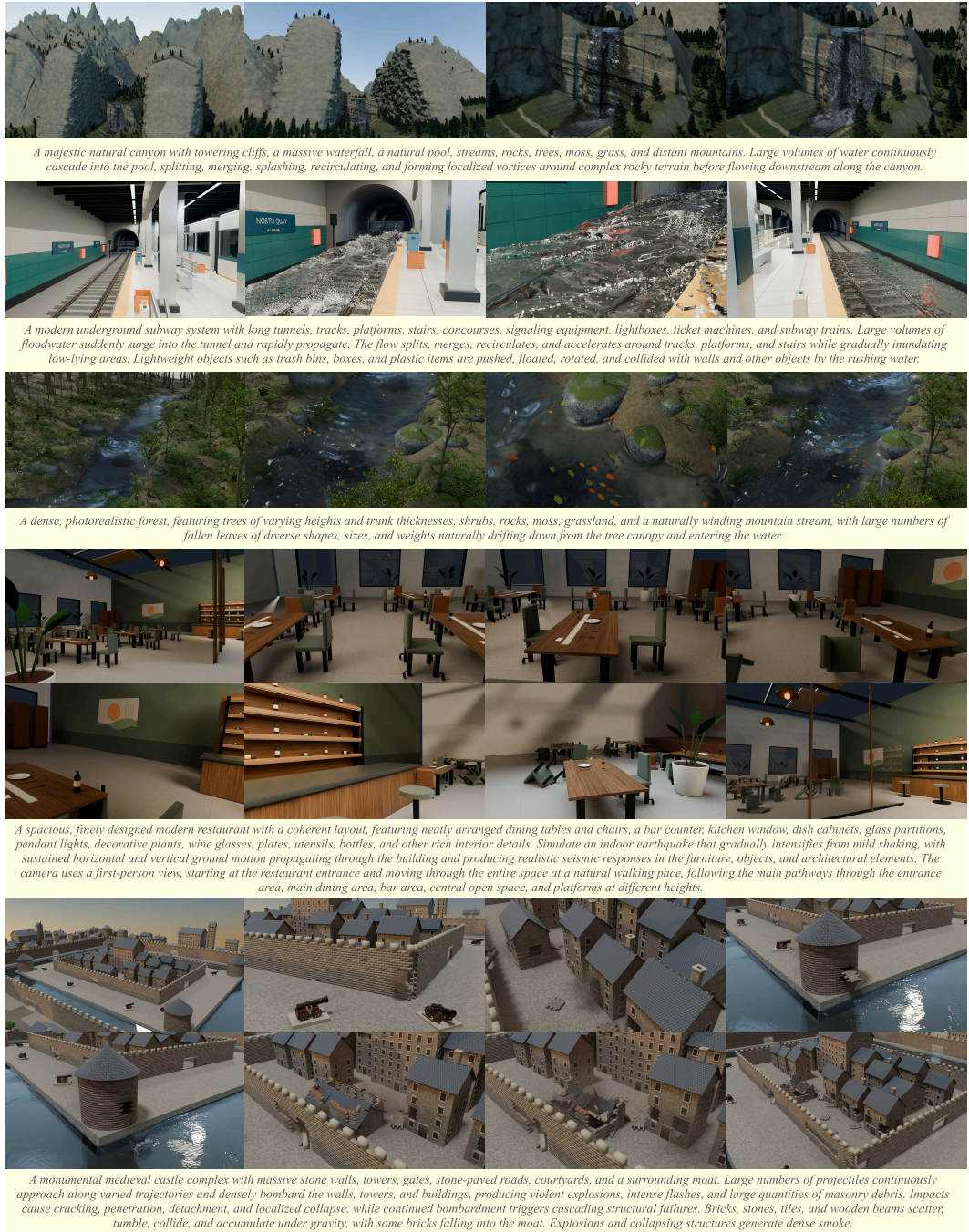}
    \caption{\textbf{Additional main results.} All results shown here are generated by WorldAgent with GPT-6 Astra as the language model backbone, using a single natural-language prompt as input without iterative user intervention.}

    \label{fig:more_main}
\end{figure*}

\begin{figure*}[t]
    \centering
    \includegraphics[width=\textwidth]{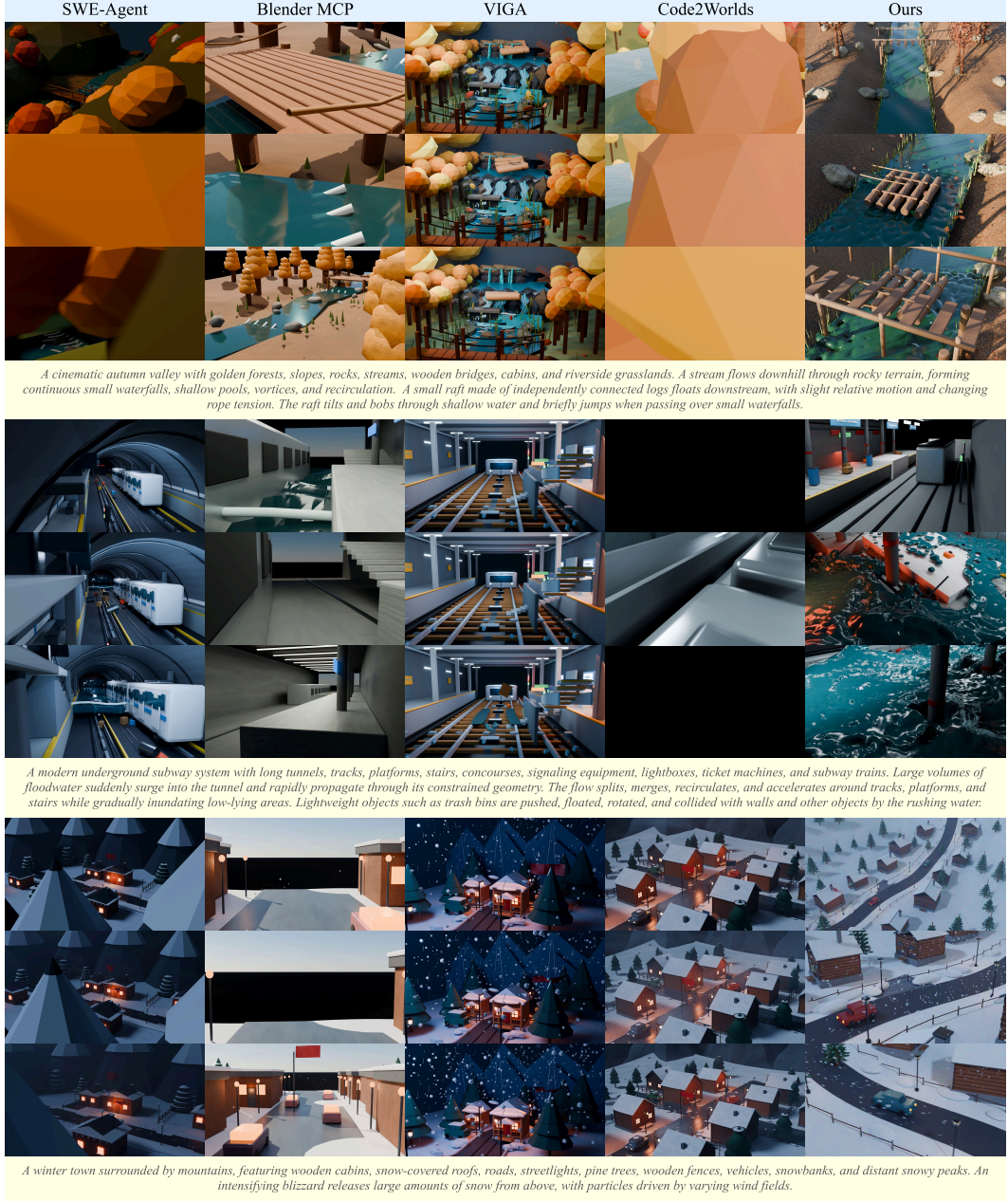}
    \caption{\textbf{Additional baseline comparisons.} All results shown here are generated by the evaluated methods using GPT-5.6 Sol as the language-model backbone, with the same single natural-language prompt as input and without iterative user intervention.}
    \label{fig:more_comparison}
\end{figure*}

\begin{figure*}[t]
    \centering
    \includegraphics[width=\textwidth]{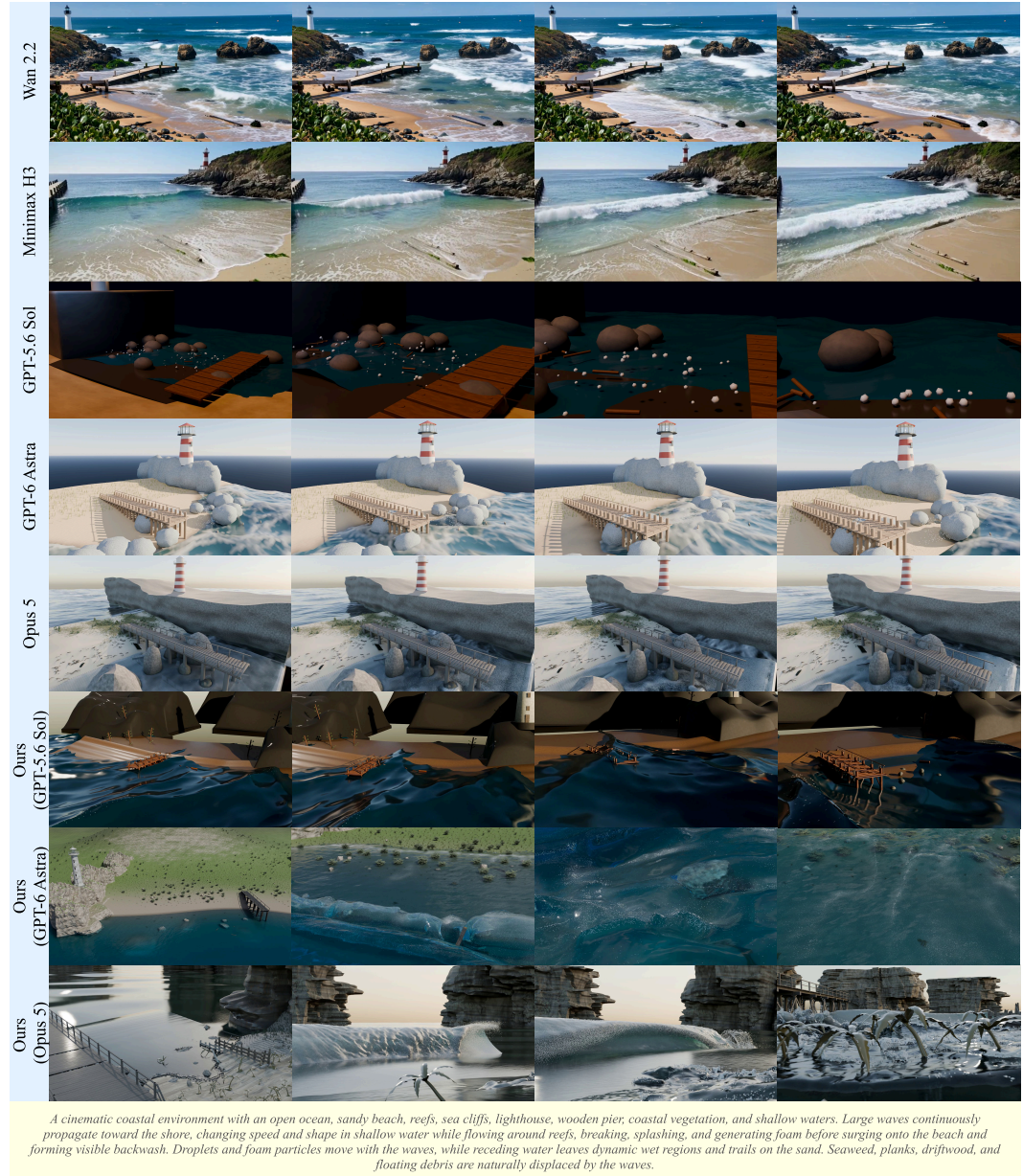}
    \caption{\textbf{Additional ablation studies.} All results shown here are generated from the same single natural-language prompt, without iterative user intervention, using different language-model backbones.
}
    \label{fig:more_comparison_model}
\end{figure*}

\begin{figure*}[t]
    \centering
    \includegraphics[width=\textwidth]{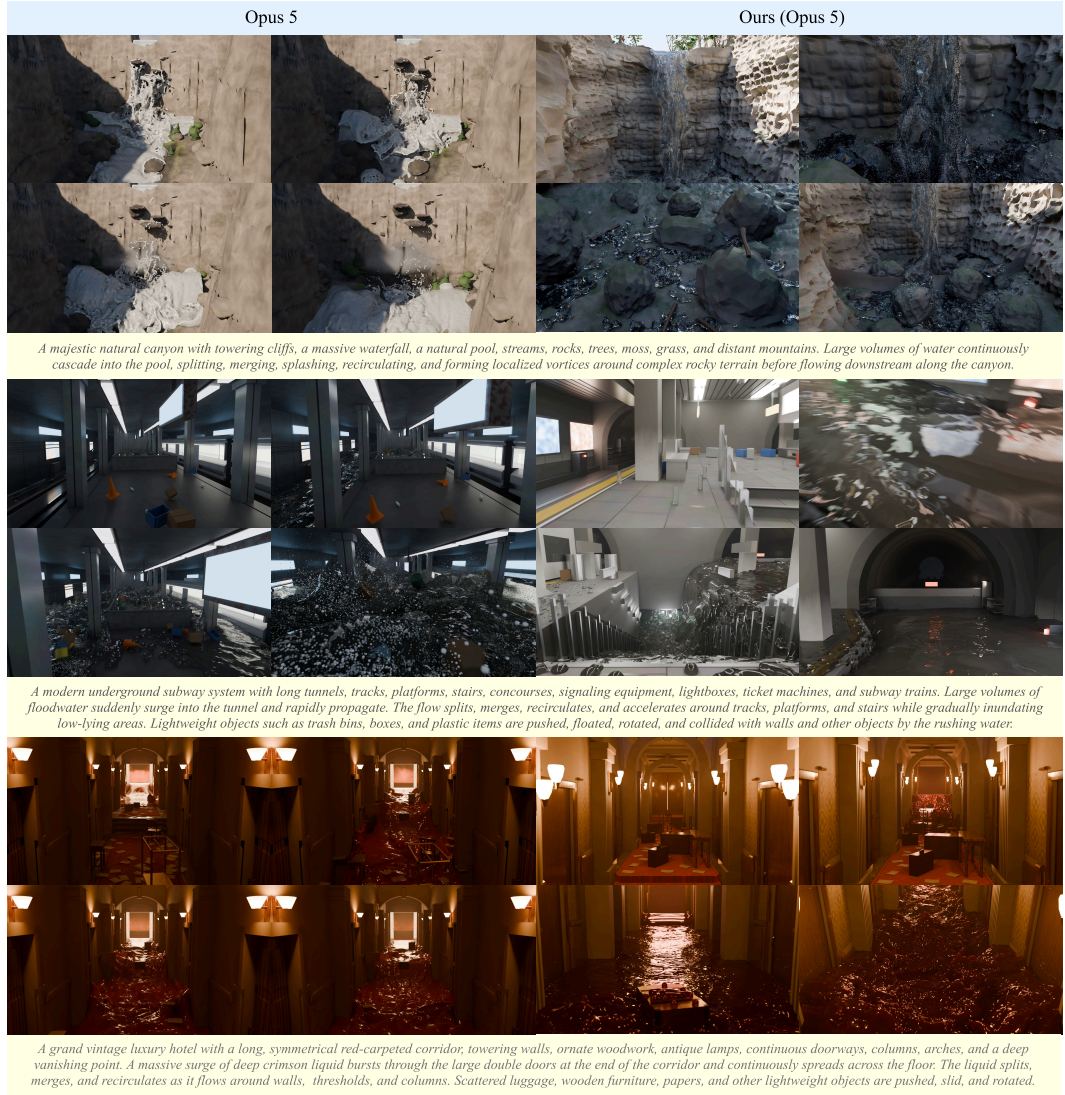}
    \caption{\textbf{Additional ablation studies.} Results shown here are generated by direct prompting with Opus 5 (Left) and by WorldAgent with Opus 5 (Right), without iterative user intervention.
}
    \label{fig:more_comparison_opus}
\end{figure*}

\end{document}